\documentclass[letterpaper, 10 pt, conference]{ieeeconf}  

\IEEEoverridecommandlockouts                              

\title{\LARGE \bf Corridor-based Adaptive Control Barrier \& Lyapunov Functions for Safe Mobile Robot Navigation}

\author{Nicholas Mohammad, Nicola Bezzo
\thanks{Nicholas Mohammad and Nicola Bezzo are with the Autonomous Mobile Robots Lab (AMR Lab) and the Department of Electrical and Computer Engineering, University of Virginia, Charlottesville, VA 22903, USA 
        {\tt\small \{nm9ur, nbezzo\}@virginia.edu}}%
}%

\newcommand{\subparagraph}{}

\usepackage{graphicx}
\usepackage{epstopdf}
\usepackage{amsmath}
\usepackage{amssymb}
\usepackage{subfigure}
\usepackage{multirow}
\usepackage{pbox}
\usepackage{algorithm}
\usepackage{algpseudocode}
\usepackage{titlesec}
\usepackage{bm}
\usepackage{url}
\usepackage{dblfloatfix} 
\usepackage{booktabs} 
\usepackage{siunitx}  
\usepackage{float}
\usepackage[colorinlistoftodos]{todonotes}

\usepackage{algpseudocode,algorithm,algorithmicx}  
\algrenewcommand\algorithmicrequire{\textbf{Precondition:}}  
\algrenewcommand\algorithmicensure{\textbf{Postcondition:}}

\titleformat{\subsubsection}[runin] 
  {\itshape\normalsize}          
  {\arabic{subsubsection})}         
  {.5em}                             
  {}                                

\titlespacing*{\subsubsection}{1em}{0.5em}{.5em}

\newcommand{\norm}[1]{\left\lVert#1\right\rVert}
\newcommand{\xihat}{\hat{\xi}}
\newcommand{\rupper}{\overline{\bm{r}}}
\newcommand{\rlower}{\underline{\bm{r}}}

\renewcommand{\baselinestretch}{.95}

\expandafter\def\expandafter\normalsize\expandafter{%
    \normalsize%
    \setlength\abovedisplayskip{3pt}%
    \setlength\belowdisplayskip{6pt}%
    \setlength\abovedisplayshortskip{-5pt}%
    \setlength\belowdisplayshortskip{2pt}%
}

\usepackage{xcolor}

\DeclareMathOperator*{\infimum}{inf}

\begin{document}

\graphicspath{ {./figs2/} }
\maketitle
\begin{abstract}
Safe navigation in unknown and cluttered environments remains a challenging problem in robotics. Model Predictive Contour Control (MPCC) has shown promise for performant obstacle avoidance by enabling precise and agile trajectory tracking, however, existing methods lack formal safety assurances. To address this issue, we propose a general Control Lyapunov Function (CLF) and Control Barrier Function (CBF) enabled MPCC framework that enforces safety constraints derived from a free-space corridor around the planned trajectory. To enhance feasibility, we dynamically adapt the CBF parameters at runtime using a Soft Actor-Critic (SAC) policy. 
The approach is validated with extensive simulations and an experiment on mobile robot navigation in unknown cluttered environments. 

\end{abstract}

\vspace{-5pt}
\section{Introduction} \label{sec:intro}
As robotic systems become more integrated into society, ensuring their stable and safe operation is becoming more critical. Consider autonomous vehicles applications, for example -- despite significant advancements in perception, planning, and control, they still experience crash rates more than twice those of human drivers \cite{chougule2024safety}. This underscores a key challenge faced by the industry: achieving reliable and safe operation across diverse environments remains an open problem.

\begin{figure}[t!]
    \centering
    \includegraphics[width=0.96\linewidth]{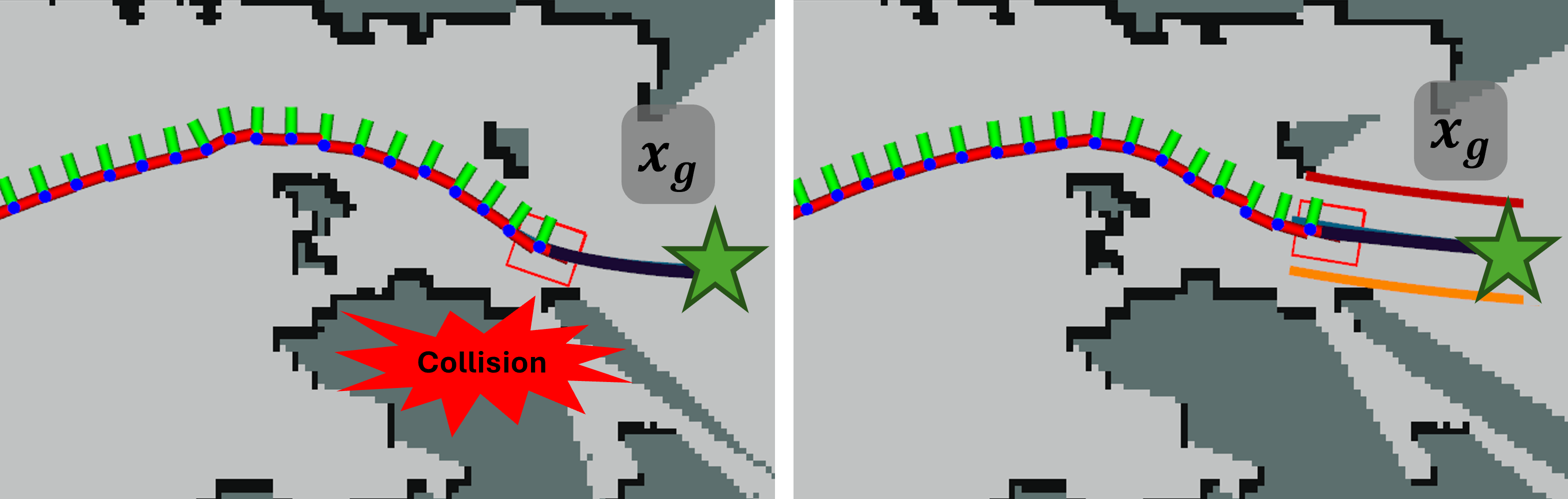}
        \vspace{-7pt}
    \caption{(Left) The high-level motion planner generates a trajectory too close to obstacles, leading to a collision. (Right) Our proposed MPCC framework incorporates corridor-based CBF constraints to ensure system safety.}
    \label{fig:intro}
    \vspace{-18pt}
\end{figure}

Traditionally, autonomous systems rely on a multi-stage motion planning pipeline that includes both a high-level, geometric path planner and low-level controller. Failures in either stage can lead to safety violations (e.g., collisions), as explored in our prior works \cite{mohammad2024gp}, \cite{mohammad2025saccbf}. Consider the scenario presented in Fig.~\ref{fig:intro}, where an autonomous system tracks receding-horizon trajectories to navigate through an unknown environment toward a goal state $\bm{x}_g$. In the case on the left, the system collides with an obstacle because the high level planner generates a trajectory that, while obstacle free, passes too close to obstacles. In contrast, as shown in the case on the right, if the system instead constructs an obstacle-free corridor (red and orange curves) around the planned path and the low level controller enforces constraints to keep the system within this safe set, safety can be ensured while still reaching the goal.


Control Barrier Functions (CBFs) provide a principled approach to enforce such safety constraints by synthesizing control inputs that prevent the system from leaving a defined safe set \cite{ames2019cbf}. However, in cluttered environments, CBFs can be overly conservative, preventing the system from passing through narrow corridors and other dense obstacle regions. To address this limitation, Control Lyapunov Functions (CLFs) are often used alongside CBFs to balance safety with stability, promoting progress toward the final goal without compromising safety \cite{liberzon1999clf}. Since both CBFs and CLFs are constraints on control inputs, they are well suited for integration within an Optimal Control Problem (OCP). Given this work's focus on trajectory tracking for obstacle avoidance, we use Model Predictive Contour Control (MPCC) \cite{lam2010mpcc} as the low-level control scheme, as it has recently demonstrated agile and precise tracking capabilities \cite{krinner2024mpcc}.

To enhance the safety and stability of MPCC, we propose a CLF-CBF formulation that constructs an obstacle-free safe set around the planned trajectory. Building on \cite{arrizabalaga2024corridor}, we generate obstacle-free corridors at runtime along the trajectory, and express the safe region as multiple CBFs, which serve as safety constraints within the MPCC. While this prevents collisions, the secondary CLF constraints cannot maintain stability when the CBF constraints are overly conservative. We have shown in our prior work \cite{mohammad2025saccbf} that CBFs can be overly conservative due to the class K function parameter, $\alpha$, which balances safety and feasibility. A low $\alpha$ value prioritizes safety but restricts the set of valid control inputs, especially in narrow corridors, while a high value increases the set of valid control inputs, risking more aggressive motion. To address this tuning problem, we propose a Soft Actor-Critic (SAC) \cite{haarnoja2019sac, mohammad2025saccbf} framework to adapt CBF parameters at runtime within the proposed MPCC formulation.


This paper presents two main contributions: 1) A free-space corridor-based CBF formulation that ensures the system remains within a safe region around the planned trajectory, and 2) A CLF-CBF enabled MPCC that balances stability and safety while tracking high-level path plans to a final goal in unknown environments, with dynamic adaptation of the CBF parameters at runtime. 

\section{Related Work} \label{sec:rel_lit}
MPCC \cite{lam2010mpcc} has recently demonstrated state-of-the-art performance in agile trajectory tracking \cite{krinner2024mpcc,   romero2022mpcc}. 
However, these MPCC formulations lack an inherent safety mechanism, making them vulnerable to collisions in unknown, cluttered environments.

To enforce safety constraints while respecting system dynamics, CBFs have emerged as a popular tool. A widely used approach builds on \cite{ames2019cbf}, where a Quadratic Program (QP) with CLF and CBF constraints is solved at runtime to synthesize safe control actions \cite{ames2014cbfqp, xu2018teleop}. However, this single-step QP is inherently myopic, ensuring safety only at the current timestep. To address this, \cite{son2019cbfcontinuous} integrate continuous-time CBF constraints into a Model Predictive Controller (MPC), with \cite{zeng2021discretecbfmpc} extending this approach to discrete-time CBFs. In the context of MPCC, safety with respect to obstacles is critical. Many CBF methods define safety constraints based on the distance to closest obstacle \cite{aali2022cbfavoidance, li2023cbf}. However, these formulations suffer from non-smooth gradients, often leading to unstable optimization problems, especially in narrow corridors. Alternative approaches enforce CBF constraints for all nearby obstacles \cite{he2022racingcbf}, significantly increasing computation. 

To address these issues, recent work has explored defining CBF constraints in an embedding that can represent the entire obstacle set compactly. \cite{narkhede2022cbfcorridor} constrain the system to remain within a corridor of discrete, convex polytopes. However, small polytope intersections can lead to situations where the system refuses to progress towards a final goal. Instead, we adopt an approach based on \cite{arrizabalaga2024corridor}, where a continuous, obstacle-free corridor is formed around the trajectory by generating two offset curves. However, the CBF formulation requires selecting a class K function parameter that balances safety and feasibility, which is difficult, if not impossible to select one appropriate value for an environment. To address this, we extend our prior Soft Actor-Critic-based adaptation framework \cite{mohammad2025saccbf}, originally designed for a single CBF constraint, to the proposed CLF-CBF-enabled MPCC, enabling dynamic adjustment of multiple CBF constraints generated by the free-space corridor formulation at runtime.

To our knowledge, the proposed approach is the first to enhance the stability and safety of the MPCC by using a CLF and free-space corridor CBF constraint, with parameter adaptation from a SAC policy to improve feasibility.

\section{Preliminaries}\label{sec:preliminaries}
In this section, we provide a brief background on the concepts of Control Lyapunov Functions (CLFs) and Control Barrier Functions (CBFs).

\subsection{Control Lyapunov Functions}
Consider the following nonlinear control-affine system:
\begin{equation}\label{eq:nonlinear_sys}
    \dot{\bm{x}} = f(\bm{x}) + g(\bm{x})\bm{u}, \quad \bm{x} = [\bm{p} \enspace \bm{\eta}]^T
\end{equation}
where $\bm{x} \in \mathcal{X} \subset \mathbb{R}^n$ consists of the position $\bm{p} = [x \enspace y]^T \in \mathbb{R}^2$ and internal states $\bm{\eta} \in \mathbb{R}^{n-2}$. The control input is given by $\bm{u} \in \mathbb{R}^m$, and the dynamics are defined by the locally Lipschitz continuous functions $f  : \mathbb{R}^n \to \mathbb{R}^n$ and $g : \mathbb{R}^n \to \mathbb{R}^{n \times m}$. To ensure stability in guiding the system towards a desired equilibrium state $\bm{x}^*$ or set $\mathcal{X}^*$, we introduce the notion of a Control Lyapunov Function (CLF), which provides a means to design a stabilizing control law.

\textbf{Definition 1 (Control Lyapunov Function \cite{liberzon1999clf}}): For the system in \eqref{eq:nonlinear_sys}, a positive definite, continuously differentiable function $V : \mathcal{X} \rightarrow \mathbb{R}$ is a \textit{Control Lyapunov Function} if $\forall \bm{x} \in \mathcal{X}$:
\begin{equation}
    \infimum_{\bm{u} \in \mathcal{U}} \Big( \underbrace{L_f V(\bm{x}) + L_g V(\bm{x}) \bm{u}}_{\dot{V}(\bm{x}, \bm{u})} + \psi_e(V(\bm{x})) \Big)  \leq 0,
\end{equation}
where $\psi_e(\cdot) \in K_\infty$ is an extended class-$K$ function, typically of the form $\psi_e(x)=\psi x$, $\psi \in \mathbb{R}^+$. $L_f V(\bm{x}) : \mathcal{X} \rightarrow \mathbb{R}$ and $L_g V(\bm{x}) : \mathcal{X} \rightarrow \mathbb{R}^m$ are the Lie derivatives of $V(\bm{x})$ with respect to $f(\bm{x})$ and $g(\bm{x})$ respectively. 

\subsection{Control Barrier Functions}
While CLFs can be used to enforce stability for driving a system to a point (or set), CBFs can be used to enforce safety by framing the problem as enforcing the invariance of a designated ``safe'' set $\mathcal{C} \subseteq \mathcal{X}$. More formally, $\mathcal{C}$ is defined as the superlevel set of a continuously differentiable function $h : \mathcal{X} \rightarrow \mathbb{R}$:
\begin{align} \label{eq:safe_set}
\mathcal{C} &= \left\{\bm{x} \in \mathcal{X} \; : \; h\left(\bm{x}\right) \geq 0\right\}  \\
\partial\mathcal{C} &= \left\{\bm{x} \in \mathcal{X} \; : \; h\left(\bm{x}\right) = 0\right\}  \\
\text{Int}\left(\mathcal{C}\right) &= \left\{\bm{x} \in \mathcal{X} \; : \; h\left(\bm{x}\right) > 0\right\},
\end{align}
where $\partial\mathcal{C}$ and $\text{Int}\left(\mathcal{C}\right)$ denote the boundary and interior of $\mathcal{C}$ respectively. To ensure safety of the system \eqref{eq:nonlinear_sys}, the safe set $\mathcal{C}$ must be \textit{forward invariant}:

\textbf{Definition 2 (Forward Invariance): } The set $\mathcal{C}$ is forward invariant if, for every initial condition $\bm{x}_0 \in \mathcal{C}$, the solution $\bm{x}_t$ of \eqref{eq:nonlinear_sys} satisfies $\bm{x}_t \in \mathcal{X}$ $\forall t \geq 0$. 

The CBF $h(\cdot)$, when defined as follows, can render the safe set $\mathcal{C}$ forward invariant, and therefore ensure safety:

\textbf{Definition 3 (Control Barrier Function \cite{ames2019cbf}}): Let $\mathcal{C} \subseteq \mathcal{X}$ define the 0-superlevel set of a continuously differentiable function $h : \mathcal{X} \rightarrow \mathbb{R}$ as defined in \eqref{eq:safe_set}, then $h(\bm{x})$ is a CBF for the system in \eqref{eq:nonlinear_sys} if $\forall \bm{x} \in \mathcal{C}$:
\begin{equation}\label{eq:cbf_const}
    \sup_{\bm{u} \in \mathcal{U}} \Big(\underbrace{L_fh(\bm{x}) + L_gh(\bm{x})\bm{u}}_{\dot{h}(\bm{x}, \bm{u})} + \alpha_e(h(\bm{x}))\Big) \geq 0,
\end{equation}
where $\alpha_e \in K_\infty$ is of the form $\alpha_e(x) = \alpha x$, $\alpha \in R^+$. 
Given a feedback control law $\bm{u}=k(\bm{x})$ for the system \eqref{eq:nonlinear_sys}, the stability and safety assurances of the CLF and CBF can be incorporated by using the following minimally perturbing CLF-CBF Quadratic Program (CLF-CBF QP) \cite{ames2019cbf}:
\begin{subequations}\label{eq:cbf_clf_qp}
\begin{align}
    u^*(\bm{x}) =
    & \enspace \underset{\bm{u}, \delta}{\text{argmin}} \quad \frac{1}{2}\norm{k(\bm{x}) - \bm{u}}^2 + p \delta^2 \\
    \text{s.t. } \quad
    & \enspace \dot{V}(\bm{x}, \bm{u}) + \psi V(\bm{x}) \leq \delta, \label{eq:clf_cons_qp} \\
    & \enspace \dot{h}(\bm{x}, \bm{u}) + \alpha h(\bm{x}) \geq 0. \label{eq:cbf_cons_qp}
\end{align}
\end{subequations}
where $\delta \in \mathbb{R}$ is a slack variable penalized by $p \in \mathbb{R}^+$, ensuring the solvability of the CLF-CBF QP. When necessary, $\delta$ is used to relax the condition on stability in order to uphold the guarantee of safety.

    
\section{Problem Formulation} \label{sec:problem}  
In this work we focus on developing a safety-focused, CLF-CBF-enabled low-level contour tracking controller $k_s(\bm{x}) : \mathcal{X} \to \mathbb{R}^m$. Let \eqref{eq:nonlinear_sys} define the dynamics for a system tasked to navigate an unknown environment. A high level motion planning policy $\Pi$ generates a trajectory $\bm{r}(\xi) = [x_r(\xi), y_r(\xi)]^T \in \mathbb{R}^2$ in receding horizon fashion, parameterized by arc-length $\xi \in [\xi_0, \xi_0 + s_r]$, where $s_r$ is the trajectory length. This contour provides a high-level path plan from the system's current state $\bm{x}_t$ to a goal $\bm{x}_g$ while avoiding a set of currently sensed obstacles $\mathcal{X}_O(t) \subset \mathcal{X}$. In this work, we define $\mathcal{X}_O(t) : \mathbb{R}^2 \to [0,1]$ as a 2D costmap of resolution $r_c \ll r_o$, where $r_o$ is the circumscribing radius of the system.

Typically, $\bm{r}(\cdot)$ is optimized to minimize the travel distance to reach $\bm{x}_g$, often resulting in trajectories that pass too close to obstacles. This leaves little margin for error, and even if the $\bm{r}(\cdot)$ is obstacle-free, collisions may still occur if $k_s(\cdot)$ cannot faithfully track $\bm{r}(\cdot)$. 

\textbf{Problem 1 - Safe Contour Control:} 
Given a receding horizon policy $\Pi$ that generates $\bm{r}(\xi)$ from the system's current state $\bm{x}_t$ to the goal $\bm{x}_g$ while avoiding the currently sensed obstacle set $\mathcal{X}_O(t)$,
\begin{equation}
    \bm{r}(\xi) = \Pi(\bm{x}_t, \bm{x}_g, \mathcal{X}_O(t)),
\end{equation}
the objective is to formulate a CLF-CBF control law $\bm{u}_t = k_s(\bm{x}_t)$, based on \eqref{eq:cbf_clf_qp}, that tracks $\bm{r}(\cdot)$ while ensuring system safety. 
As shown in \cite{mohammad2025saccbf}, the effectiveness of the CBF constraint \eqref{eq:cbf_const} requires precise tuning of the $\alpha$ parameter, and typically requires adaptation as the system navigates toward $\bm{x}_g$, leading to the following problem.

\textbf{Problem 2 - Adaptive Safe Contour Control:}
Find an adaptation policy $\pi_\sigma(\cdot | \bm{s}_t)$ that dynamically adjusts $\alpha$ in real time based on the environment and system state. This policy should use a state embedding $\bm{s}_t$ representing the obstacle set $\mathcal{X}_O(t)$, system state $\bm{x}_t$, and nominal input $\bm{u}_t$, to ensure safe navigation toward $\bm{x}_g$. 

In the following sections, we discuss in detail our formulation of $k_s(\bm{x})$ as a CLF-CBF-enabled Model Predictive Contour Controller (MPCC), as well as our SAC-based policy $\pi_\sigma$ for adapting $\alpha$ at runtime. 

\section{Approach}\label{sec:approach}
\begin{figure}[th!]
    \centering
    \includegraphics[width=0.98\linewidth]{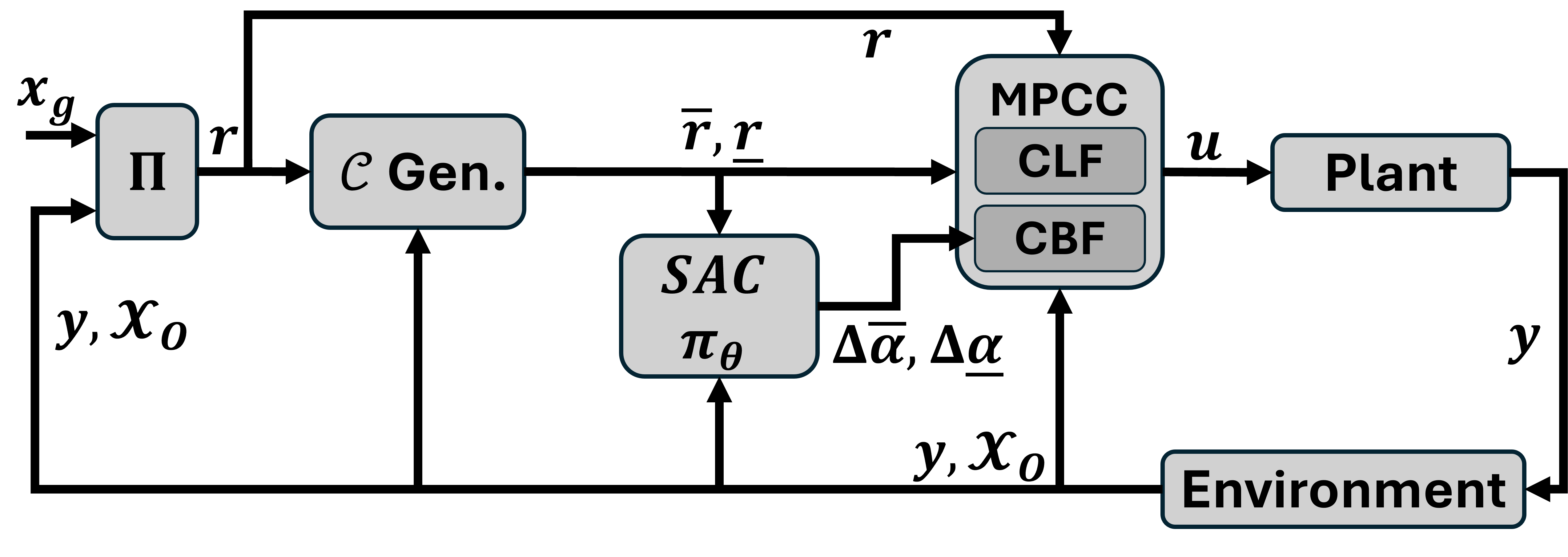}
               \vspace{-10pt}
    \caption{Block diagram of our proposed approach.}
    \label{fig:block_diagram}
           \vspace{-5pt}
\end{figure}

We propose a framework that augments the MPCC control paradigm with a free-space, corridor-based CLF-CBF scheme. This approach improves both the safety and stability of MPCC in cluttered, unknown environments while eliminating the need for manual tuning of the CBF $\alpha$ parameter. The proposed control system architecture is illustrated in Fig.~\ref{fig:block_diagram}. 
           \vspace{-5pt}
A receding horizon, high level motion-planning policy $\Pi$ generates an obstacle-free, arc-length parameterized trajectory $\bm{r}(\xi)$ from the current state $\bm{x}_t$ to the goal $\bm{x}_g$ using the sensed costmap $\mathcal{X}_O(t)$. This trajectory is passed to the corridor generation module \cite{arrizabalaga2024corridor}, which constructs a free-space corridor $\mathcal{C}$ bounded by two offset curves, $\rupper(\xi), \rlower(\xi) \in \mathbb{R}^2$. These serve as CBF constraints \eqref{eq:cbf_cons_qp} within the MPCC, with tuning parameters $\overline{\alpha}$ and $\underline{\alpha}$. 

However, overly conservative CBF constraints can lead to large slack variables $\delta$ in \eqref{eq:clf_cons_qp}, sacrificing stability for safety when $\overline{\alpha}$ or $\underline{\alpha}$ are too low. To prevent this, we extend our prior work \cite{mohammad2025saccbf} by training a SAC-based policy $\pi_\sigma$ offline to adapt $\overline{\alpha}$ and $\underline{\alpha}$ dynamically based on $\bm{x}_t$ and $\mathcal{C}$. This adaptation dynamically adjusts the CBF constraints, preventing excessive $\delta$ values -- especially in narrow corridors, where rigid constraints can hinder tracking progress. The following sections detail our framework, starting with the CLF-CBF-enabled MPCC formulation.


\subsection{Model Predictive Contour Control}\label{sec:MPCC}
Consider the discrete dynamics of the system in \eqref{eq:nonlinear_sys}:
\begin{equation}\label{eq:discrete_sys}
    \bm{x}_{k+1} = f_d(\bm{x}_k) + g_d(\bm{x}_k) \bm{u}_k, \; \bm{x}_k = [\bm{p}_k \enspace \bm{\eta}_k]^T
\end{equation}
where $\bm{x}_k \in \mathbb{R}^n$ denotes the state of the system at timestep $t_k = t_0 + k \Delta t$ and $f_d : \mathcal{X} \rightarrow \mathbb{R}^n$ and $g_d : \mathcal{X} \rightarrow \mathbb{R}^{n \times m}$ define the discrete-time dynamics of the system. 

In MPCC \cite{lam2010mpcc}, the objective is to use an Optimal Control Problem (OCP) to steer the system \eqref{eq:discrete_sys} along an arc length parameterized reference path $\bm{r}(\xi)$ while maximizing speed and minimizing tracking error. 

\textbf{Assumption 1: } The trajectory $\bm{r}(\cdot)$ is an obstacle-free, $C^2$-continuous curve parameterized by arc length such that $\frac{ds}{d\xi} = 1$, where $s$ denotes the distance traveled along the path. Note that, although arc length parameterization of general curves is nontrivial, effective techniques for parameterizing splines exist in the literature, i.e., the \textit{bisection method} \cite{wang2003parameterization}. 

\begin{figure}[th!]
    \centering
    \includegraphics[width=0.8\linewidth]{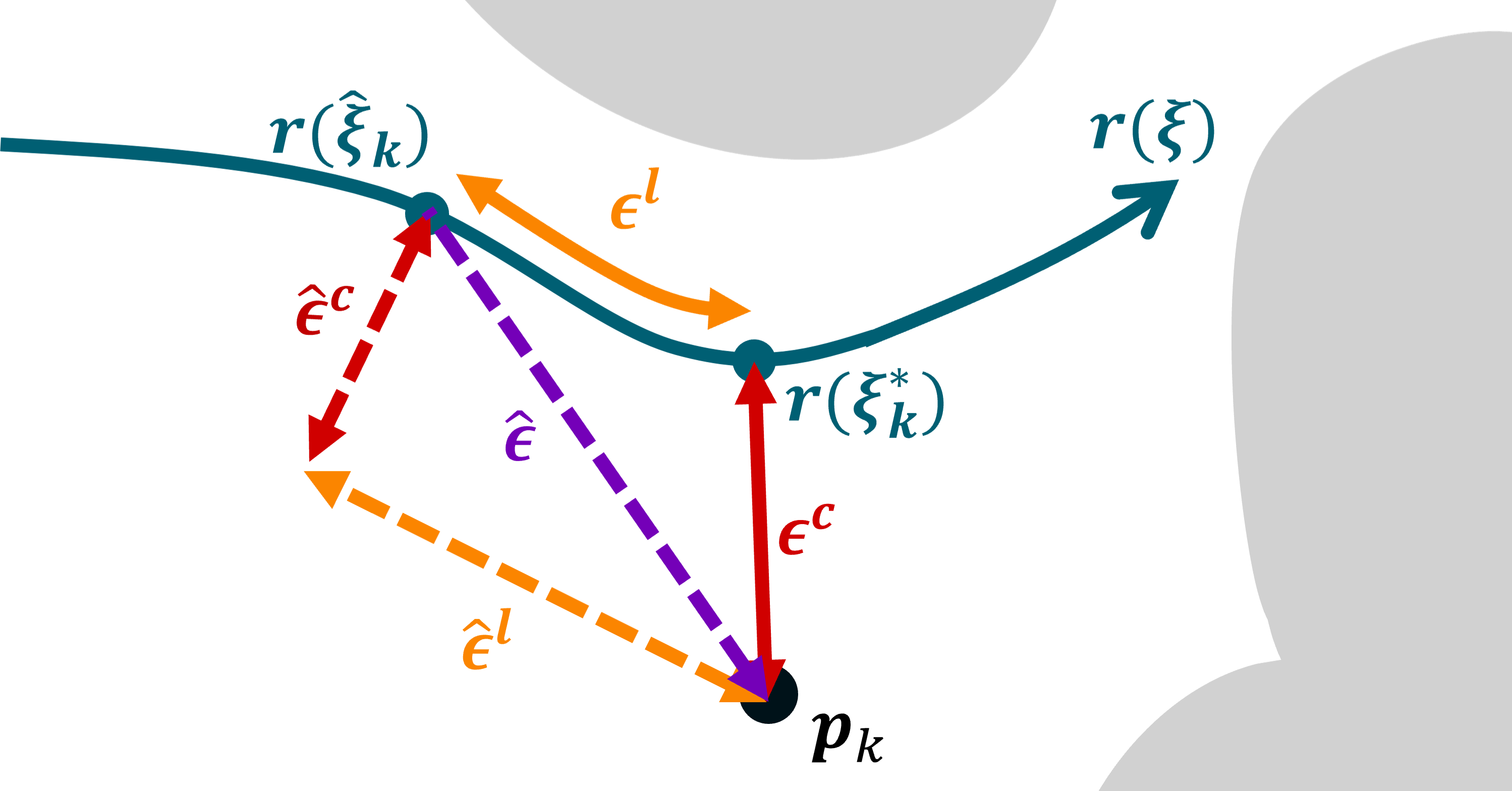}
    \caption{Illustration of MPCC contour and lag error terms.}
    \label{fig:mpcc_illustration}
       \vspace{-5pt}
\end{figure}

The tracking error minimized by the MPCC is referred to as the \textit{contour error}, $\epsilon^c \in \mathbb{R}_{\geq 0}$, defined as the normal distance of $\bm{p}_k$ to $\bm{r}(\xi)$:
\begin{equation}
    \epsilon^c = \min_\xi \norm{\bm{p}_k - \bm{r}(\xi)}_2,
\end{equation}
where the minimizing term of this optimization problem is denoted as $\xi^*_k$ (see Fig.~\ref{fig:mpcc_illustration}).
       \vspace{-5pt}

However, using this definition of $\epsilon^c(\cdot)$ in an online OCP is infeasible, because it is itself an optimization problem. To approximate $\xi^*_k$ in real time, a virtual state term $\xihat_k \in [0, s_r]$ is introduced, which has the following dynamics:
\begin{equation}
    \hat{\xi}_{k+1} = \xihat_k + \nu_k \Delta t
\end{equation}
where $\nu_k = \frac{\Delta\hat{\xi}}{\Delta t} \in [0, \nu^+]$, $\nu^+ \in \mathbb{R}^+$ is a virtual input that represents the speed of progression along the path, and is determined by the OCP. The approximated total position error at $k$ is then given by $\hat{\bm{\epsilon}}(\bm{p}_k, \xihat_k)=\bm{p}_k - \bm{r}(\xihat_k)$. The approximate contour error, $\hat{\epsilon}^c$ (dashed red in Fig.~\ref{fig:mpcc_illustration}), can then be expressed as the normal component of $\hat{\bm{\epsilon}}(\bm{p}_k, \xihat_k)$:

\begin{equation}
    \hat{\epsilon}^c(\bm{p}_k, \xihat_k) = \norm{\bm{P}_\perp(\xihat_k)\hat{\bm{\epsilon}}(\bm{p}_k, \xihat_k)}_2,
\end{equation}
where $\bm{P}_\perp(\xihat_k) = (\bm{I}_2-\bm{t}_r(\xihat_k)\bm{t}_r^T(\xihat_k))$ is the normal projection matrix, $\bm{I}_2 \in \mathbb{R}^{2\times 2}$ is the identity matrix and $\bm{t}_r(\xihat_k)=\frac{d\bm{r}(\xi)}{d\xi}\big|_{\xi=\xihat_k}$ denotes the unit tangent vector to the curve. The accuracy of this approximation has been shown in \cite{lam2010mpcc} to be directly linked to the \textit{lag error} $\epsilon^l = ||\xihat_k - \xi^*_k||_2$. Since this term also depends on $\xi^*_k$, it is approximated as $\hat{\epsilon}^l(\bm{p}_k, \xihat_k)$ (dashed orange in Fig.~\ref{fig:mpcc_illustration}), which is the scalar projection of $\hat{\bm{\epsilon}}(\cdot)$ onto $\bm{t}_r(\xihat_k)$:
\begin{equation}
    \hat{\epsilon}^l(\bm{p}_k, \xihat_k) = \bm{t}_r^T(\xihat_k)\hat{\bm{\epsilon}}(\bm{p}_k, \xihat_k).
\end{equation}

With the error terms and virtual states defined, the MPCC problem for the system in \eqref{eq:discrete_sys} with CLF and CBF constraints can be formulated as an OCP. A direct formulation would treat $\xihat$ as a virtual state and $\nu_k$ as a virtual input. However, this would result in no limits on how quickly $\nu_k$ can change, leading to noisy control inputs. To avoid this, we follow standard MPCC implementations \cite{krinner2024mpcc} and use the progress acceleration $\dot{\nu}$ as a virtual input instead. Throughout the rest of the paper, the system state $\bm{x}_k=[\bm{p}_k \enspace \bm{\eta}_k \enspace \xihat_k \enspace \nu_k]^T$ implicitly includes the virtual MPCC states, and the input vector $\bm{u}$ includes $\dot{\nu}$. In the following formulation, we abbreviate $\hat{\epsilon}^c(\bm{p}_k, \xihat_k)$ as $\hat{\epsilon}^c_k$, and similarly for $\hat{\epsilon}^l_k$:
\begin{subequations}\label{eq:mpcc_ocp}
\begin{align}
    & \underset{\bm{x}, \bm{u}, \delta}{\text{argmin}}
    & & \sum_{k=1} ^N q_c (\hat{\epsilon}^c_k)^2 + q_l (\hat{\epsilon}^l_k)^2 - q_\nu \nu_k + p\delta^2 \label{eq:mpcc_cost} \\
    & \text{subject to}
    & & \bm{x}_0 = \bm{x} \\
    & & & \bm{x}_{k+1} = f_d(\bm{x}_k) + g_d(\bm{x}_k)\bm{u}_k \\
    & & & \hat{\xi}_{k+1} = \xihat_k + \nu_k \Delta t \\
    & & & \nu_{k+1} = \nu_k + \dot{\nu}_k \Delta t \\
    & & & \dot{V}(\bm{x}_k, \bm{u}_k) + \psi V(\bm{x}_k) \leq \delta \label{eq:mpcc_clf_const} \\
    & & & \dot{h}(\bm{x}_k, \bm{u}_k) + \alpha h(\bm{x}_k) \geq 0 \label{eq:mpcc_cbf_const} \\
    & & & \bm{x}_k \in \mathcal{X}, \enspace \xihat_k \in [0, s_r], \enspace \nu_k \in [0, \nu_k^+] \\
    & & & \bm{u}_k \in \mathcal{U}, \enspace \dot{\nu}_k \in [0, \dot{\nu}^+].
\end{align}
\end{subequations}
The objective of the OCP is to minimize the position error $\hat{\bm{\epsilon}}(\cdot)$ while maximizing progress ($\nu_k)$ along $\bm{r}(\cdot)$, with user-defined weights $q_c, q_l, q_\nu, p \in \mathbb{R}^+$ balancing each term. 

\textbf{Assumption 2:} The OCP used for the MPCC is capable of maintaining $\hat{\epsilon}^l_k \approx 0$ if $q_l \gg q_c$. As shown in \cite{lam2010mpcc}, if $\hat{\epsilon}^l_k \approx 0$, then $\xihat_k \approx \xi^*_k$ and $\hat{\epsilon}^c_k$, and $\hat{\epsilon}^l_k$ converge to their true values. 

\subsection{CLF Formulation for MPCC}
With the MPCC defined, a Lyapunov function is introduced to enforce system stability as the system in \eqref{eq:discrete_sys} tracks the trajectory $\bm{r}(\xihat_k)$. In this work, we define stability to be the equilibrium point at which the total tracking error of the MPCC is zero, i.e. $\hat{\epsilon}^c_k = 0$ and $\hat{\epsilon}^l_k = 0$. Using this equilibrium point in the Lyapunov formulation also assists in satisfaction of \textit{Assumption 2} mentioned above. The full Lyapunov candidate function $V(\bm{x}_k)$ is defined as:
\begin{align}
    V(\bm{x}_k) &= \lambda_c (\hat{e}^c_k)^2 + \lambda_l (\hat{e}^l_k)^2
\end{align}
where $\lambda_c, \lambda_l \in \mathbb{R}^+$ are weighting parameters associated with each error term. 

\textbf{Remark 1:} While the MPCC formulation in \eqref{eq:mpcc_ocp} aims to ensure stability through the CLF constraint in \eqref{eq:mpcc_clf_const}, the slack variable $\delta$ ensures that safety remains the priority. If enforcing stability conflicts with maintaining safety, the stability constraint may be violated to avoid unsafe behavior. The level of trade-off between stability and safety can be adjusted via the parameter $p$ in \eqref{eq:mpcc_cost}. 

\subsection{CBF Formulation for MPCC}
With the stability constraint defined, we now construct the safe set $\mathcal{C}$ and subsequent CBF formulation for our contour controller, which takes the form of an obstacle-free safety corridor around the trajectory $\bm{r}(\cdot)$.

\subsubsection{Construction of $\mathcal{C}$:}
Constructing $\mathcal{C}$ over the entire length of $\bm{r}(\cdot)$ at runtime may be computationally intractable -- especially when the total arc length $s_r$ is large. To address this, we instead generate $\mathcal{C}$ in a receding horizon fashion up to a user-specified maximum length $s^+$.

\begin{figure}[th!]
    \centering
     \subfigure[]{\includegraphics[width=0.4\linewidth]{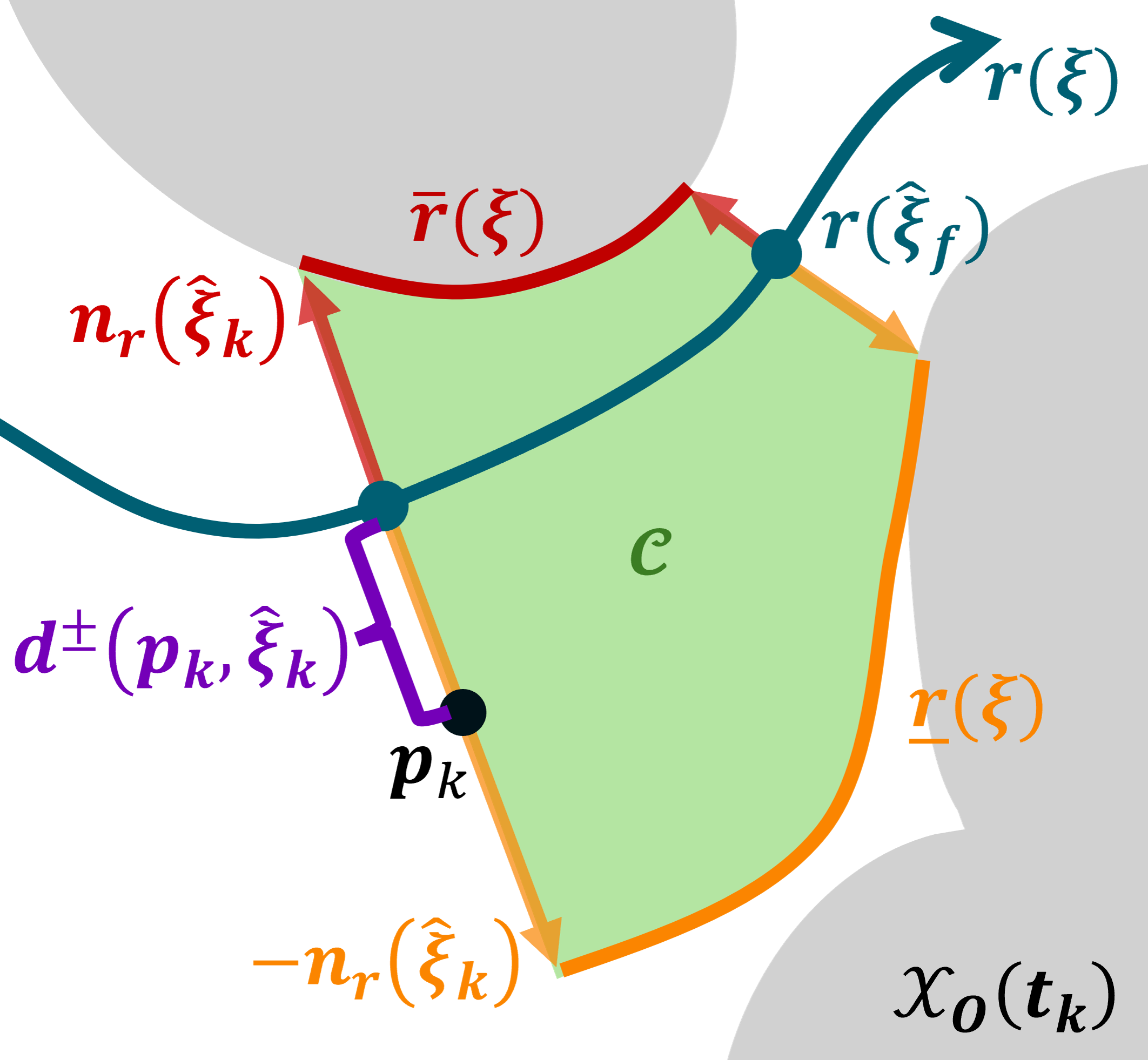}\label{fig:corridor_illustration}}
     \subfigure[]{\includegraphics[width=0.4\linewidth]{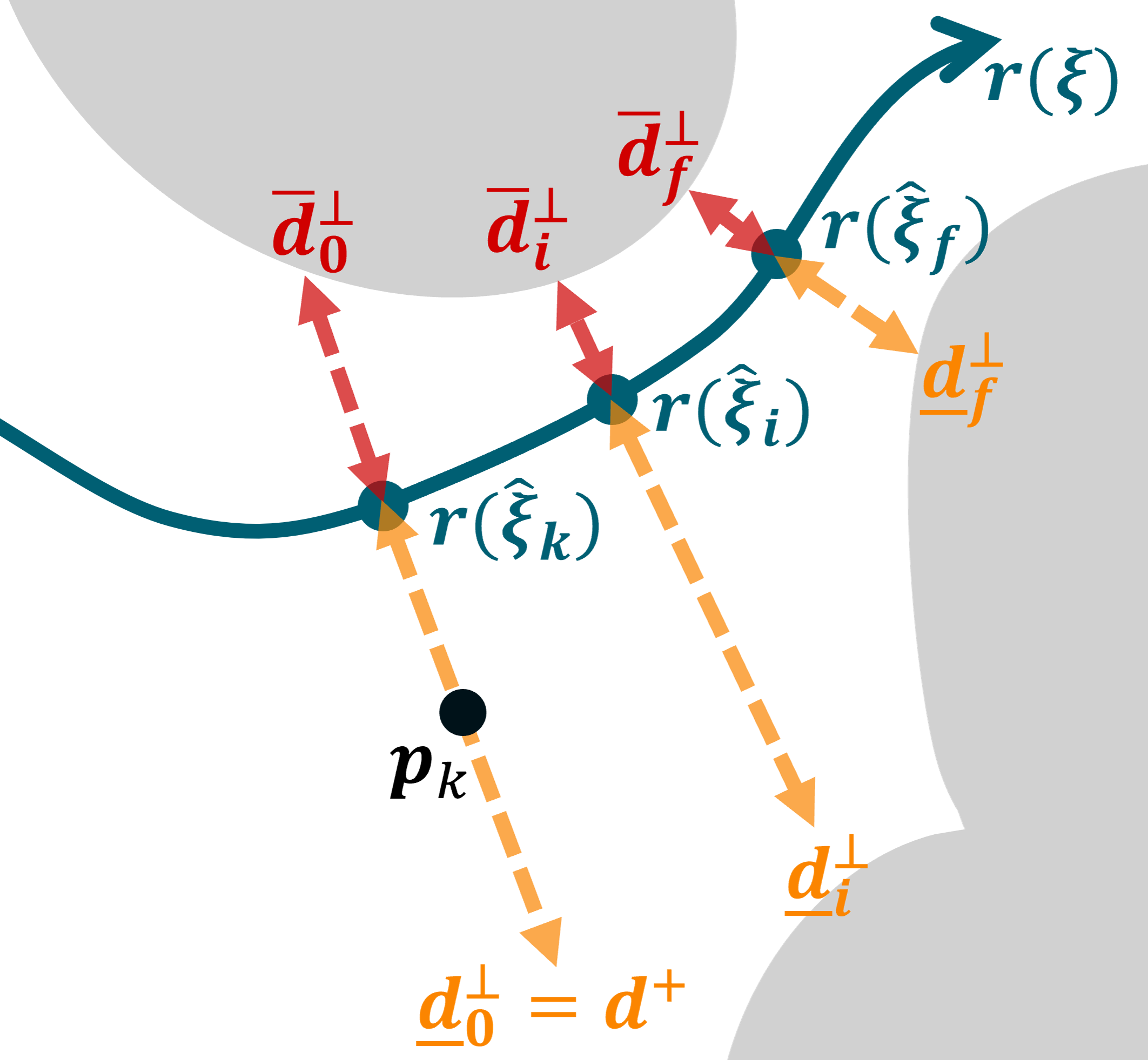}\label{fig:corridor_gen}}
    \caption{(a) Visualization of the safe set $\mathcal{C}$. (b) Illustration of the construction process for $\mathcal{C}$.}
    \label{fig:corridor}
    \vspace{-10pt}
\end{figure}

Specifically, we define $\mathcal{C}$ as the region of space between two offset curves $\rupper(\xi)$ and $\rlower(\xi)$, parameterized on the interval $\mathcal{I}_\xi = [\xihat_k, \xihat_f]$, $\xihat_f = \min\{\xihat_k  + s^+, s_r\}$ (see Fig.~\ref{fig:corridor_illustration}):
\begin{align}
    \rupper(\xi) &= \bm{r}(\xi) + \overline{d}(\xi)\bm{n}_r(\xi) \\
    \rlower(\xi) &= \bm{r}(\xi) - \underline{d}(\xi)\bm{n}_r(\xi),
\end{align}
where $\bm{n}_r(\xi) = \bm{R}(\pi/2)\bm{t}_r(\xihat)$ is the unit counter-clockwise normal, and $\bm{R}(\theta) \in \mathbb{R}^{2\times2}$ is the 2D rotation matrix. 

To construct these offset curves, we define two functions $\overline{d}(\cdot), \underline{d}(\cdot) \in \mathbb{R}_{\geq 0}$, which map the distances from $\bm{r}(\xihat)$ to the nearest obstacles above and below, respectively. However, obtaining closed-form expressions for these functions at runtime is challenging -- if not impossible -- due to the complexity of arbitrary obstacle geometries. Instead, we approximate them by sampling $N_r$ normal distances $\{\overline{d}^\perp_i\}_{i=0}^{N_r}$ and $\{\underline{d}^\perp_i\}_{i=0}^{N_r}$ to the trajectory $\bm{r}(\cdot)$ at equidistant arc lengths $\Delta\xi$ along the interval $\mathcal{I}_\xi$. In Sec.~\ref{sec:theoretical}, we provide theoretical insight on the selection of $\Delta\xi$. 

As illustrated in Fig.~\ref{fig:corridor_gen}, at each sampled point $\bm{r}(\xihat_i)$, we cast a ray of maximum distance $d^+$ along the counter-clockwise normal $\bm{n}_r(\xihat_i)$ to compute $\overline{d}^\perp_i$, the normal distance to the nearest obstacle above $\bm{r}(\xihat_i)$ in the costmap $\mathcal{X}_O(t_k)$. Similarly, we determine $\underline{d}^\perp_i$ by raycasting along $-\bm{n}_r(\xihat_i)$. 
To approximate smooth functions from the sampled distances, we model $\overline{d}(\xihat ; \overline{\bm{C}})$ and $\underline{d}(\xihat ; \underline{\bm{C}})$ as polynomials of degree $D$, parameterized by their coefficient matrices $\overline{\bm{C}}, \underline{\bm{C}} \in \mathbb{R}^{2\times (D + 1)}$:
\begin{equation}
    \overline{d}(\xihat; \overline{\bm{C}}) = \sum_{j = 0}^D \overline{\bm{C}}_j \xihat^j, \enspace \underline{d}(\xihat; \underline{\bm{C}}) = \sum_{j = 0}^D \underline{\bm{C}}_j \xihat^j,
\end{equation}
where $\overline{\bm{C}}_j, \underline{\bm{C}}_j \in \mathbb{R}^2$ denote the $j$th column of the coefficient matrices. To determine $\overline{\bm{C}}$ and $\underline{\bm{C}}$, we use a Linear Program (LP) fitting technique based on \cite{arrizabalaga2024corridor}, which aims to maximize the area between the two polynomials subject to the sampled distance constraints.


\subsubsection{CBF Formulation: }
To formally define $\mathcal{C}$, we introduce the signed normal distance $\hat{d}^\pm(\bm{p}, \hat{\xi})$ from a position $\bm{p}$ to trajectory $\bm{r}(\cdot)$ (purple in Fig.~\ref{fig:corridor_illustration}):
\begin{equation}
    \hat{d}^\pm(\bm{p}, \xihat) = \hat{\epsilon}^c(\bm{p}, \xihat)\bm{n}_r(\xihat)
\end{equation}
Intuitively, $\hat{d}^\pm(\cdot)$ represents the signed contour error $\hat{\epsilon}^c(\cdot)$. It is positive when aligned with $\bm{n}_r$ (counter-clockwise from $\bm{t}_r$), and negative otherwise. Since safety depends on keeping $\hat{d}^\pm(\bm{p}, \hat{\xi})$ within set bounds, we define the safe set $\mathcal{C}$ for a given trajectory $\bm{r}(\cdot)$ as the set of all states $\bm{x}$ where $\hat{d}^\pm(\bm{p}, \xihat)$ remains within the bounds defined by $\underline{d}(\xihat)$ and $\overline{d}(\xihat)$:
\begin{equation}
    \mathcal{C} = \left\{\bm{x} \in \mathcal{X} \enspace | \enspace \hat{d}^\pm(\bm{p}, \hat{\xi}) \in [-\underline{d}(\hat{\xi}), \overline{d}(\hat{\xi})] \right\}.
\end{equation}
Since the definition of $h(\cdot)$ depends on the dynamics of the target system, we present general candidate CBFs, $\overline{h}(\bm{x}), \underline{h}(\bm{x})$ for the relative degree $1$ case here: 
\begin{align}
    \overline{h}(\bm{x}) &= \overline{d}(\xihat) - \hat{d}^\pm(\bm{p}, \xihat) \label{eq:cbf_abv} \\
    \underline{h}(\bm{x}) &= \underline{d}(\xihat) + \hat{d}^\pm(\bm{p}, \xihat). \label{eq:cbf_blw}
\end{align}

With the CBF functions defined, the constraint in \eqref{eq:mpcc_cbf_const} in the MPCC formulation \eqref{eq:mpcc_ocp} can be replaced with the following two constraints:
\begin{align}
    \dot{\overline{h}}(\bm{x}_k, \bm{u}_k) + \overline{\alpha} \cdot \overline{h}(\bm{x}_k) \geq 0 \label{eq:corridor_cbf_abv} \\
    \dot{\underline{h}}(\bm{x}_k, \bm{u}_k) + \underline{\alpha} \cdot \underline{h}(\bm{x}_k) \geq 0 \label{eq:corridor_cbf_blw}.
\end{align}

Note that the CBF formulations depend on $\xihat$, and thus depend on satisfying \textit{Assumption 2} where $\hat{\epsilon}^l(\bm{p}, \xihat) \approx 0$. 

\textbf{Example (Unicycle Model): } To demonstrate how the corridor CBFs can be applied to a specific system, we use the unicycle model as an example, whose dynamics describe a wide variety of mobile platforms. The state for the unicycle model is $\bm{x} = [\bm{p} \enspace \theta \enspace v]^T$, consisting of SE$(2)$ pose $[\bm{p} \enspace \theta] \in \mathbb{R}^2 \times [-\pi, \pi]$ and velocity $v \in \mathbb{R}$. The MPCC augmented dynamics are given by:
\begin{equation}\label{eq:unicycle_dynamics}
    \dot{\bm{x}} =
    \begin{bmatrix}
        \dot{x} \\
        \dot{y} \\
        \dot{\theta} \\
        \dot{v} \\
        \dot{\xihat} \\
        \dot{\nu}
    \end{bmatrix}
    =
    \underbrace{
    \begin{bmatrix}
        v \cos(\theta) \\
        v \sin(\theta) \\
        0 \\
        0 \\
        \nu \\
        0
    \end{bmatrix}
    }_{f(\bm{x})}
    + 
    \underbrace{
    \begin{bmatrix}
        0 & 0 & 0 \\
        0 & 0 & 0 \\
        0 & 1 & 0 \\
        1 & 0 & 0 \\
        0 & 0 & 0 \\
        0 & 0 & 1
    \end{bmatrix}
    }_{g(\bm{x})}
    \begin{bmatrix}
        a \\
        \omega \\
        \dot{\nu}
    \end{bmatrix},
\end{equation}
where $\bm{u}=[a \enspace \omega \enspace \dot{\nu}]^T$ consists of linear acceleration $a$, angular velocity $\omega$, and progress acceleration $\dot{\nu}$.

One challenge with using the unicycle model in \eqref{eq:unicycle_dynamics} is that the CBFs defined in \eqref{eq:cbf_abv}-\eqref{eq:cbf_blw} have relative degree $k=2$ with respect to acceleration $a$ and angular velocity $\omega$. As a result, $L_gh(\bm{x})=\bm{0}$, which implies that $[a \enspace \omega]^T$ cannot be directly regulated by \eqref{eq:corridor_cbf_abv}-\eqref{eq:corridor_cbf_blw}. To address this, the following CBFs, adapted from \cite{li2023cbf} are used:
\begin{align}
    \overline{h}(\bm{x}) &= \left[\overline{d}(\xihat) - \hat{d}^\pm(\bm{p},\xihat)-r_o\right]\exp{\left\{-\bm{n}_r(\xihat) \cdot \bm{e}_{\theta} - v\lambda\right\}} \label{eq:h_func_abv_uni}, \\
    \underline{h}(\bm{x}) &= \left[\underline{d}(\xihat) + \hat{d}^\pm(\bm{p},\xihat) - r_o\right]\exp{\left\{\bm{n}_r(\xihat) \cdot \bm{e}_{\theta} - v\lambda\right\}} \label{eq:h_func_blw_uni},
\end{align}
where $r_o$ is the circumscribing radius of the system, $\bm{e}_\theta=[\cos\theta \enspace \sin\theta]$, and $\lambda \in \mathbb{R}^+$ is a user-defined parameter ensuring $v\lambda \in (0,1)$. The relative degree of the proposed CBFs with respect to $a$ and $\omega$ is reduced to $1$, where $L_g\overline{h}(\bm{x}) = \overline{h}(\bm{x}) [-\lambda \enspace -\bm{t}_r(\xihat)\bm{e}_\theta \enspace 0]$. The expression for $L_g\underline{h}(\bm{x})$ is nearly identical, but is omitted here for brevity. 

\textbf{Remark 2:} Although the CBFs in \eqref{eq:h_func_abv_uni}-\eqref{eq:h_func_blw_uni} do not explicitly regulate $\dot{\nu}$ (third entry in $L_g\overline{h}(\bm{x})$), the MPCC formulation is designed such that $\hat{\epsilon}^l \approx 0$ (Assumption 2). Since the CBFs already regulate linear acceleration $a$ and $\hat{\epsilon}^l \approx 0$, $\dot{\nu}$ is implicitly regulated.

With the CBFs defined, in the following section, we discuss in detail our SAC framework for adapting the $\underline{\alpha}$ and $\overline{\alpha}$ parameters at runtime to ensure robust and agile navigation.

\subsection{SAC for CBF Adaptation}
So far we have presented the CBF formulation with a constant $\alpha$ parameter, but in this work, $\alpha_k$ varies over time and is adapted by a learned policy $\pi_\sigma$. This policy selects actions $\bm{a}_k$ to adjust $\overline{\alpha}$ and $\underline{\alpha}$ within a user defined interval $[\alpha^-, \alpha^+]$ at runtime: 
\begin{equation}
    \bm{\alpha}_{k+1} = \bm{\alpha}_0 + \sum_{i=1}^k \bm{a}_i,
\end{equation}
where $\bm{a}_k=[\Delta\overline{\alpha}_k, \Delta\underline{\alpha}_k]$ defines the adjustments, $\bm{\alpha}_k=[\overline{\alpha}_k, \underline{\alpha}_k]$, and $\bm{\alpha}_0$ are the initial values for the parameters. Sec.~\ref{sec:simulation} discusses the selection of $\bm{\alpha}_0$ and $[\alpha^-, \alpha^+]$ further. 

This adaptation process can be modeled as a Markov Decision Process (MDP), $\mathcal{T}=(\mathcal{S}, \mathcal{A}, R, \mathcal{P}, \rho_0)$, where $\mathcal{S}$ and $\mathcal{A}$ define valid states and actions, and the reward function $R : \mathcal{S} \times \mathcal{A} \times \mathcal{S} \to \mathbb{R}$ assigns a reward for transitioning to state $\bm{s}_{k+1}$ after taking action $\bm{a}_k$ at state $\bm{s}_k$. The transition probability $\mathcal{P} : \mathcal{S} \times \mathcal{A} \times \mathcal{S} \rightarrow [0,1]$, defines the probability 
of transitioning to state $\bm{s}_{k+1}$, given current state $\bm{s}_k$ and action $\bm{a}_k$. Lastly, $\rho_0$ is the probability distribution over initial states. 

In this work, the policy $\pi_\sigma$ is trained using the Soft Actor-Critic (SAC) algorithm \cite{haarnoja2019sac}, which maximizes an objective function balancing expected cumulative reward and entropy:
\begin{equation}\label{eq:sac_return}
    J_\pi = \mathop{\mathbb{E}}_{\bm{\tau} \sim \pi}\left[\sum_{k=0}^\infty \gamma^k \Big( R(\bm{s}_k, \bm{a}_k, \bm{s}_{k+1}) + \beta H(\pi_\sigma)\Big)\right],
\end{equation}
where $\gamma \in (0, 1]$ is a discount factor on future rewards $R(\cdot)$, $\bm{\tau} = \left(\bm{s}_0, \bm{a}_0, \bm{s}_1, \bm{a}_1, \dots \right)$ is a sequence of states $\bm{s}_k$ and actions $\bm{a}_k$ in the environment, $H(P)=\mathbb{E}_{x \sim P}[-\log P(x)]$ is the entropy term, and $\beta$ is a temperature parameter that controls the trade-off between exploration and exploitation, which is automatically tuned during training \cite{haarnoja2019sac}.

To train the model, observed state transitions and associated rewards are stored in a replay buffer $\mathcal{D}$ as tuples $\bm{\zeta}_k = (\bm{s}_{k}, \bm{a}_{k}, r_{k}, \bm{s}_{k+1}, d_k)$, where $r_k = R(\bm{s}_k, \bm{a}_k, \bm{s}_{k+1})$ and $d_k$ indicates terminal states (e.g., collisions or goal reached). The data from this buffer is then used to train the policy model $\pi_\sigma$ and critic functions $Q_{\phi_1}$ and $Q_{\phi_2}$, parameterized by $\bm{\sigma}, \bm{\phi_1}$, and $\bm{\phi}_2$, respectively, using stochastic gradient to minimize the following loss functions:
\begin{align}\label{eq:sac_loss_functions}
    L(\bm{\sigma}, \mathcal{D}) &= \mathop{\mathbb{E}}_{\substack{\bm{s_k}\sim\mathcal{D} \\ \bm{a}_k\sim\pi_\sigma}} \left[-\min_{j=1,2} Q_{\phi_j}(\bm{s}_k,\bm{a}_k) + \beta\log\pi_\sigma(\bm{a}_k|\bm{s}_k) \right], \nonumber \\
    L(\bm{\phi}_i, \mathcal{D}) &= \mathop{\mathbb{E}}_{\bm{\zeta}_k \sim \mathcal{D}} \left[\left( Q_{\phi_i}(\bm{s}_k, \bm{a}_k)-y(r_k,\bm{s}_{k+1},d_k) \right)^2 \right],
\end{align}
where $y(\cdot)$ is the target for the critic functions and is computed from the received reward and target value of the next state-action pair \cite{haarnoja2019sac}. Training is implemented following the process in \cite{mohammad2025saccbf}.
\begin{figure*}[b!]
    \centering
     \subfigure[]{\includegraphics[width=0.453\linewidth]{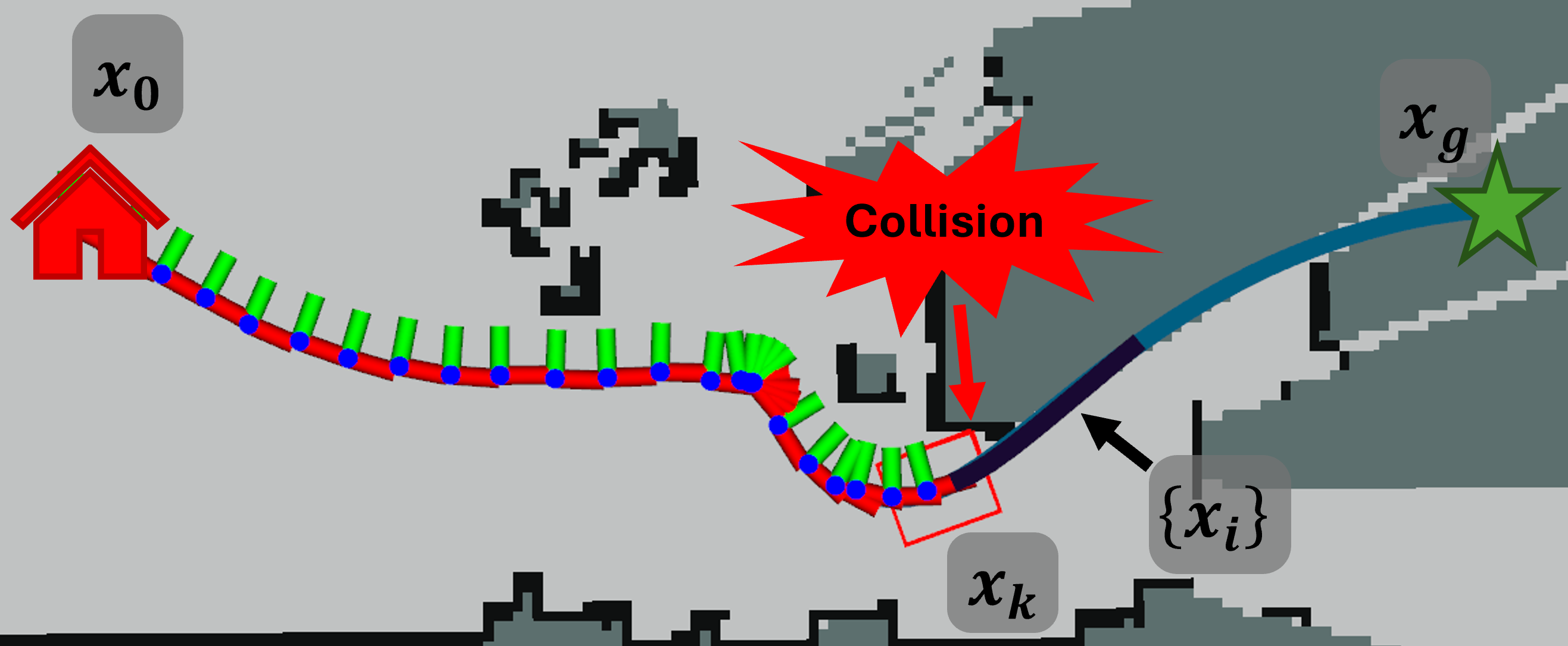}\label{fig:no_cbf_sim_result}}
     \subfigure[]{\includegraphics[width=0.46\linewidth]{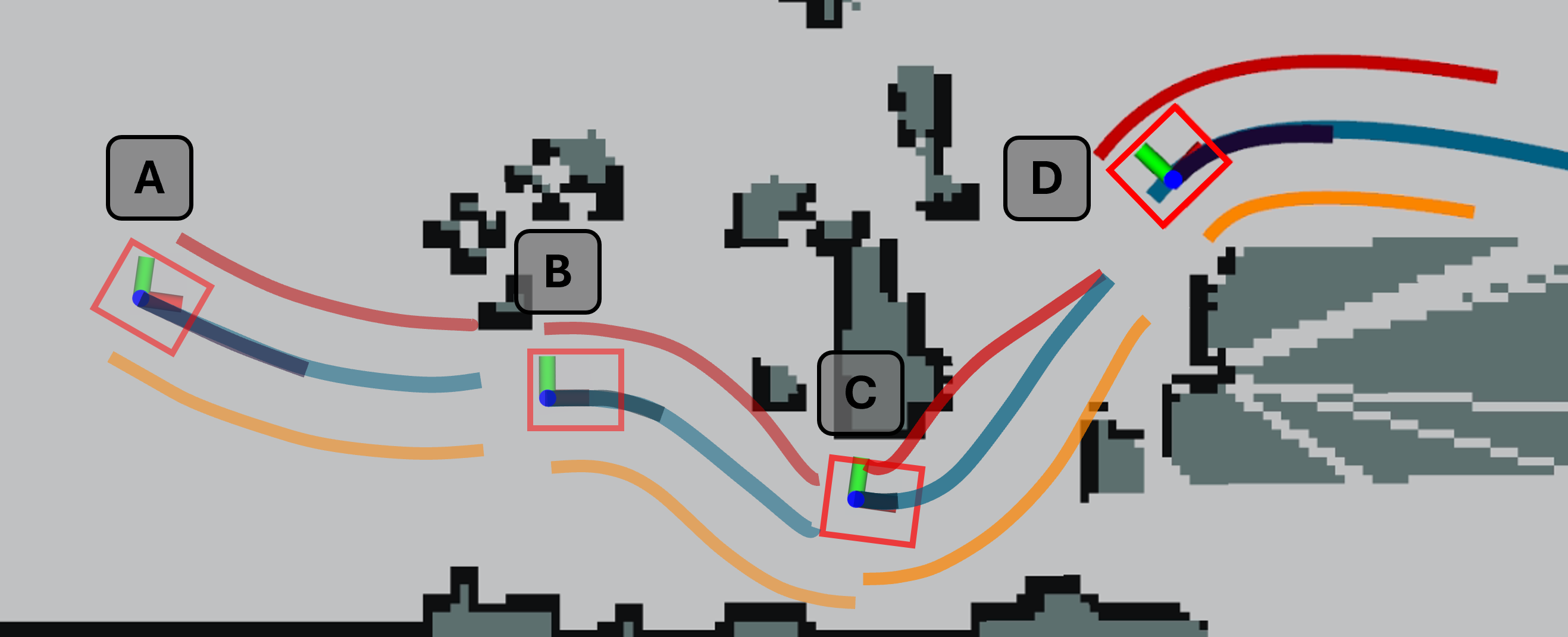}\label{fig:sac_cbf_sim_result}} \\
     \subfigure[]{\includegraphics[width=0.32\linewidth]{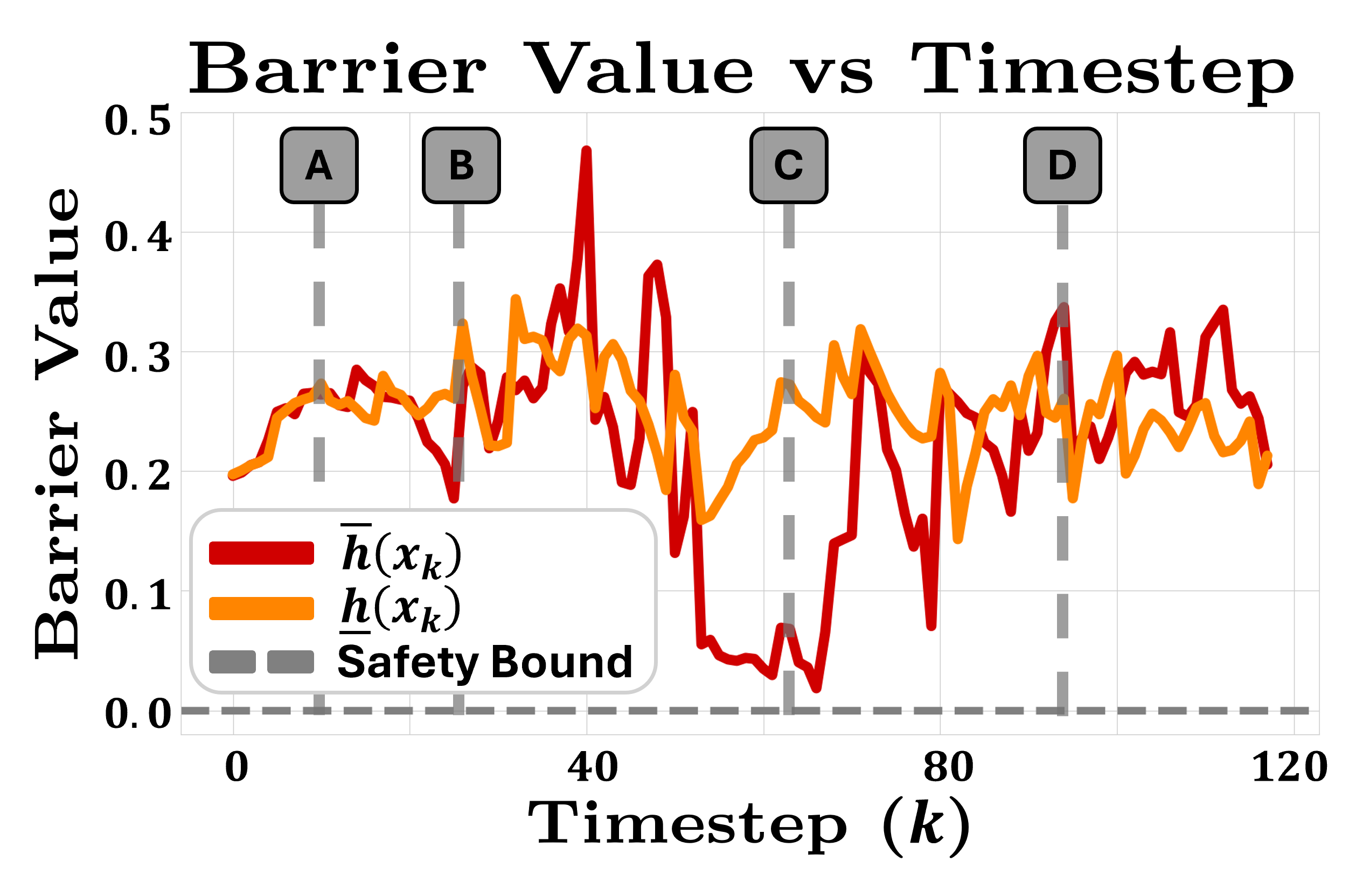}\label{fig:h_value_plt}} 
     \subfigure[]{\includegraphics[width=0.32\linewidth]{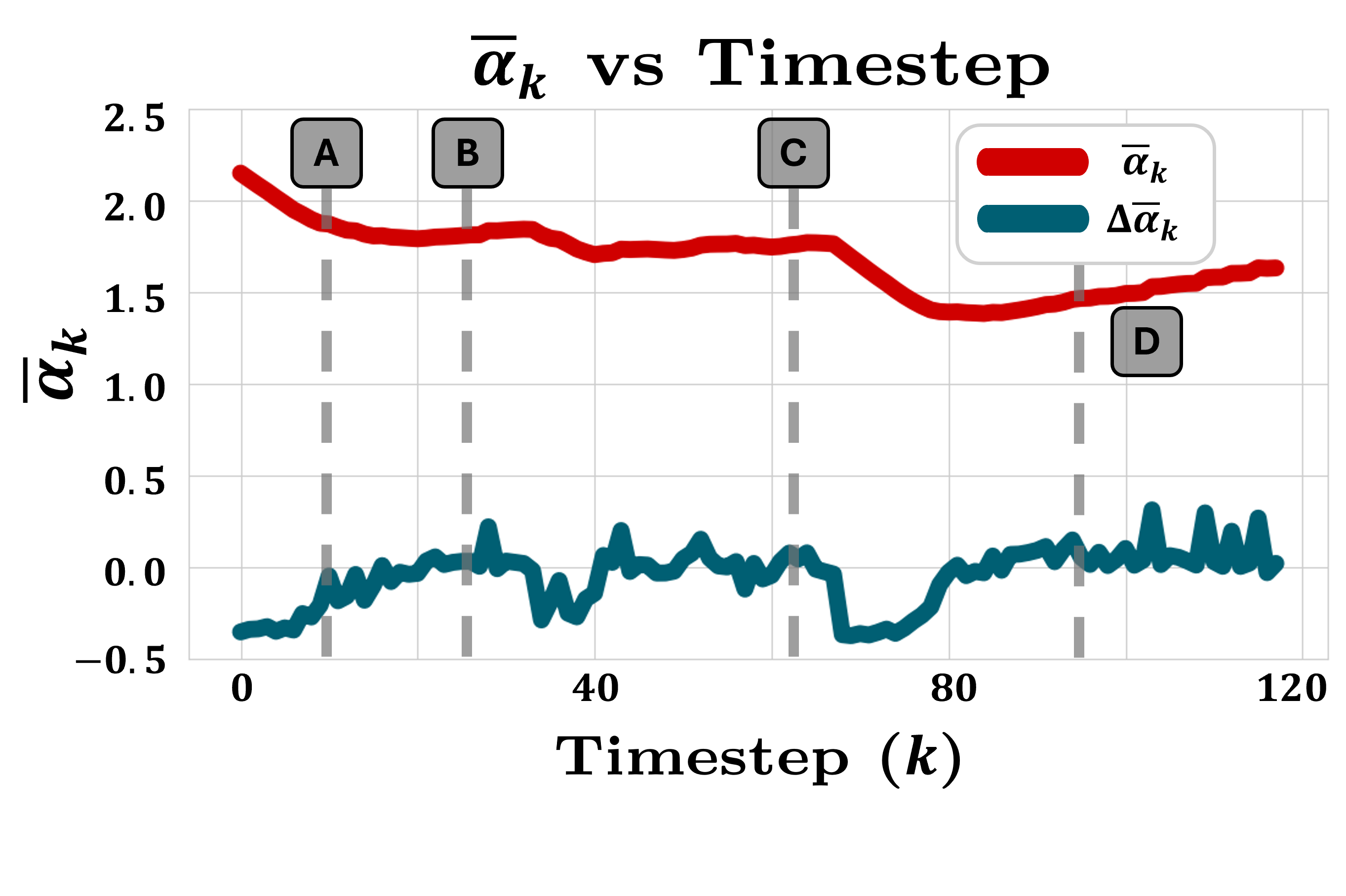}\label{fig:alpha_abv_plt}}
     \subfigure[]{\includegraphics[width=0.32\linewidth]{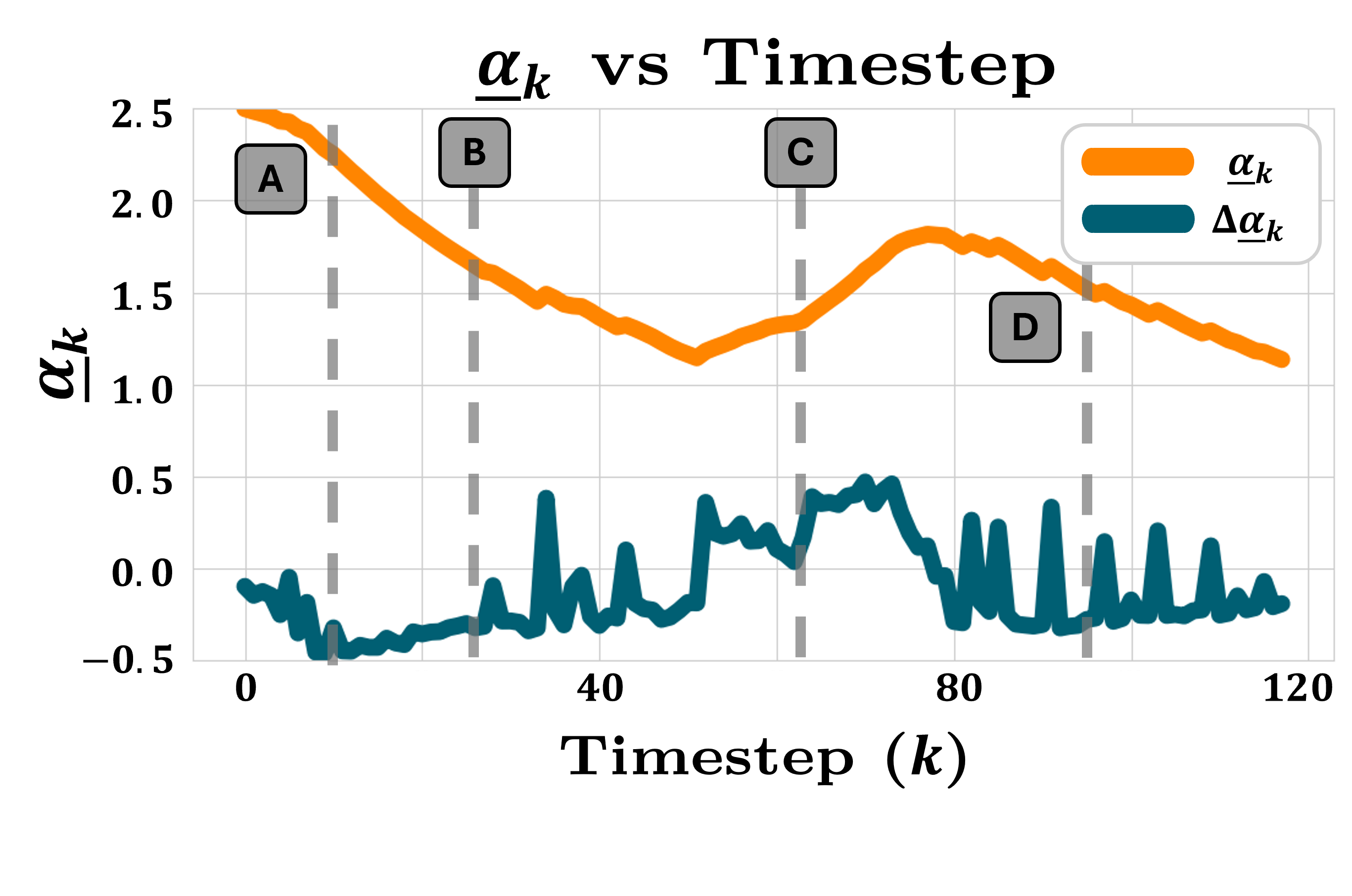}\label{fig:alpha_blw_plt}}
    \caption{Simulation results using (a) MPCC-BASE and (b) MPCC-CBF-SAC. (c) CBF values over the course of the navigation task for MPCC-CBF-SAC shown in (b). (d) and (e) show plots of $\overline{\alpha}_k$ and $\underline{\alpha}_k$ and adaptation values $\Delta\overline{\alpha}_k$ and $\Delta\underline{\alpha}_k$ for (b).}
    \label{fig:simulation_results}
\end{figure*}

\subsection{SAC State and Reward Construction}
To ensure the SAC policy adapts the $\alpha$ parameters without compromising the feasibility of the QP in \eqref{eq:mpcc_ocp}, both the reward function $R(\bm{s}_k, \bm{a}_k, \bm{s}_{k+1})$ and the SAC state $\bm{s}_k$ must be carefully designed. We begin by first defining the reward function as follows:
\begin{equation}\label{eq:reward_fun}
    R(\cdot) = -\gamma_\nu(1-\frac{\nu_k}{\nu^+}) - \gamma_b b(\bm{\alpha}_k, \bm{a}_k) - \gamma_h c(\bm{h}_k) - \gamma_z Z_k,
\end{equation}
where $\gamma_v, \gamma_b, \gamma_h, \gamma_z \in \mathbb{R}^+$ are weighting parameters, $\bm{h}_k = [\overline{h}(\bm{x}_k,\xihat_k), \underline{h}(\bm{x}_k, \xihat_k)]$ represents the CBF values, and $\bm{\alpha}_k = [\overline{\alpha}_k, \underline{\alpha}_k]$ denotes the CBF constraint parameters. The binary variable $Z_k \in \{0,1\}$ is set to $1$ if the MPCC failed to converge to a solution in the previous iteration.

The first term incentivizes the policy to maximize the progress velocity $\nu_k$, promoting agile motion along the trajectory. The function $b(\cdot)$ imposes a quadratic penalty when $\overline{\alpha}_k$ and $\underline{\alpha}_k$ deviate from the midpoint of the permissible range, $\alpha_m=\frac{\alpha^+-\alpha^-}{2}$, and applies a large fixed penalty if either parameter exceeds the interval $[\alpha^-,\alpha^+]$. The function $c(\cdot)$ imposes a large negative reward if either of the CBFs in \eqref{eq:cbf_abv}-\eqref{eq:cbf_blw} become negative, enforcing safety. Finally, the $Z_k$ term discourages actions that lead to solver infeasibility, penalizing policies that result in solver failures. 

With the reward function defined, the state vector $\bm{s}_k$ is constructed to include all relevant parameters required for the SAC policy $\pi_\sigma(\cdot | \bm{s}_k)$ to dynamically adjusts $\bm{\alpha}_k$:
\begin{equation}
    \bm{s}_k = \left[\bm{\Gamma}(\bm{\eta}_k) \enspace \nu_k \enspace \bm{d}_k \enspace \bm{h}_k(\xihat_k) \enspace \bm{\alpha}_k \enspace Z_k \right],
\end{equation}
where $\bm{d}(\xihat_k)=[\overline{d}(\xihat_k), \underline{d}(\xihat_k)]$, and $\bm{\Gamma}(\cdot)$ is a vector-valued function which maps relevant internal states $\bm{\eta}_k$ to $\bm{s}_k$. For the unicycle model in \eqref{eq:unicycle_dynamics}, the internal state mapping simplifies to $\bm{\Gamma}(\bm{\eta}) = \bm{\eta} = [\theta \enspace v]$.  Although $\bm{d}$ and $\bm{\Gamma}$ are not explicitly expressed in the reward function \eqref{eq:reward_fun}, they are provided to the policy for additional context, since they are both needed to compute the CBFs \eqref{eq:h_func_abv_uni}-\eqref{eq:h_func_blw_uni}. Note that we exclude $\bm{p}_k$ from $\bm{s}_k$ in order to remove any dependence of the policy on absolute position values. 

\subsection{Theoretical Analysis}\label{sec:theoretical}
Ensuring that the constructed corridor $\mathcal{C}$ remains obstacle-free is a necessary condition for guaranteeing system safety. Another necessary condition is recursive feasibility, however, since it is not generally ensured for Model Predictive Control (MPC) or other nonlinear optimization problems \cite{zeng2021discretecbfmpc}, we instead derive necessary conditions under which $\mathcal{C}$ is guaranteed to be free of undetected obstacles. 

The sampling resolution for generating $\rupper(\cdot)$ and $\rlower(\cdot)$ must be fine enough to prevent large gaps between consecutive points, $\bm{p}_i$ and $\bm{p}_{i+1}$, that could allow grid cells of $\mathcal{X}_O(t)$ to protrude into the safe set. Thus, the step-size of $\Delta\xi$ must be chosen carefully to ensure that no gaps exceeding the costmap resolution, $r_c$, exist between consecutive samples. 


\textbf{Lemma 1:} To ensure that the maximum distance between consecutive sampled points satisfies $\norm{\bm{p}_i - \bm{p}_{i+1}} \leq r_c$, the step-size $\Delta \xi$ along $\bm{r}(\cdot)$ must satisfy: 
\begin{equation}\label{eq:sampling_condition}
    \Delta\xi \leq \frac{r_c}{1 + d^+\kappa^+}
\end{equation}

\begin{proof}
Consider two consecutive points obtained through the raycast sampling procedure:
\begin{align*}
\bm{p}_i &= \bm{r}(\xi_i) + \bm{n}_r(\xi_{i}) d \\
\bm{p}_{i+1} &= \bm{r}(\xi_{i+1}) + \bm{n}_r(\xi_{i+1}) d
\end{align*}

The goal is to bound the maximum distance between consecutive samples. Using the triangle inequality, a bound on the difference can be established:
\begin{align*}
\norm{\bm{p}_{i+1} - \bm{p}_{i}} &=  \lVert\bm{r}(\xi_{i+1})-\bm{r}(\xi_{i}) - \\
& \quad \, d(\bm{n}_r(\xi_{i+1}) - \bm{n}_r(\xi_i))\rVert \\
& \leq  \norm{\Delta\bm{r}} + d\norm{\Delta \bm{n}_r}
\end{align*}

Using the Frenet-Serret frame for $\bm{r}(\cdot)$, first order approximations for $\Delta \bm{r}=\bm{r}(\xi_{i+1})-\bm{r}(\xi_{i})$ and $\Delta \bm{n}=\bm{n}_r(\xi_{i+1})-\bm{n}_r(\xi_{i})$ can be derived from the following relations when $\Delta\xi \ll s_r$: $\Delta\bm{r}/\Delta\xi \approx \bm{t}_r(\xi_i)$ and $\Delta\bm{n}_r/\Delta\xi \approx -\kappa(\xi_i) \bm{t}_r(\xi_i)$. Using these approximations along with the fact that $\norm{\bm{t}_r(\xi)}=1$ by \textit{Assumption 2}, we can derive the following:
\begin{align*}
\norm{\bm{p}_{i+1} - \bm{p}_{i}} & \leq \norm{\Delta\bm{r}} + d\norm{\Delta \bm{n}_r} \\
& \approx \norm{\bm{t}_r(\xi_i)\Delta\xi} + d\norm{\kappa(\xi_i)\Delta\bm{n}_r} \\
& \leq \Delta\xi + d^+ \kappa^+ \Delta \xi \\
& = (1+d^+\kappa^+)\Delta\xi
\end{align*}

Therefore, to ensure the sampling condition is met, the following constraint must be satisfied: $\norm{\bm{p}_{i+1} - \bm{p}_{i}} \leq (1+d^+\kappa^+)\Delta\xi \leq r_c$. Solving for $\Delta\xi$ in terms of $r_c$ directly leads to \eqref{eq:sampling_condition}.
\end{proof}

This result ensures that the discretization does not introduce gaps larger than $r_c$, which could allow undetected obstacles to protrude into $\mathcal{C}$. However, even with a sufficiently dense sampling for offset curve generation, obstacles with dimension larger than $r_c$ can still protrude into $\mathcal{C}$, leading to the following lemma.

\textbf{Lemma 2:} Consider the costmap $\mathcal{X}_O(t_k)$ at timestep $k$, with resolution $r_c$. If the maximum distance between two consecutive raycasted samples $\bm{p}_i, \bm{p}_{i+1}$ is at most $r_c$, then the maximum distance that an undetectable obstacle can protrude into $\mathcal{C}$ is at most $\frac{r_c}{2}$.

\begin{proof}
    Consider a protruding costmap cell with point $\bm{p}_o$ protruding into $\mathcal{C}$, located between two consecutive raycasted samples $\bm{p}_i$ and $\bm{p}_{i+1}$. The three points form a triangle denoted as $\triangle(\bm{p}_i, \bm{p}_{i+1}, \bm{p}_o)$, with base width $b=\norm{\bm{p}_{i+1} - \bm{p}_{i}}$ and unknown height $x$, which represents the protrusion distance into the safe corridor $\mathcal{C}$. 
    The area of the triangle can be expressed in two equivalent ways $A(\triangle(\cdot))=\frac{1}{2}bx=\frac{1}{2}b^2\cos(\theta) \sin(\theta)$. Rearranging to solve for $x$ produces the following $x=b\frac{\sin(2\theta)}{2}$. Given that $b \leq r_c$, the expression is maximized at $x=r_c/2$ with $b=r_c$ and $\theta=\pi/4$. 
\end{proof}

\textbf{Corollary 1:} If $\rupper(\cdot)$ and $\rlower(\cdot)$ are shifted away from the obstacles by at least a distance of $\frac{r_c}{2}$, then it is guaranteed that no obstacles with a dimension larger than $r_c$ are within $\mathcal{C}$. The proof follows directly from \textit{Lemma 2}.

Since we shift the offset curves by the system's circumscribing radius $r_o$ in \eqref{eq:h_func_abv_uni}-\eqref{eq:h_func_blw_uni}, and traditional costmaps have much finer resolution than $r_o$, $\rupper(\cdot)$ and $\rlower(\cdot)$ are guaranteed to be obstacle-free. In the following section, we demonstrate these theoretical results and provide empirical evidence for the feasibility constraints defined by $\mathcal{C}$ and $\pi_\sigma$. 


\section{Results}\label{sec:simulation}

\subsection{Simulation}
The proposed framework was trained and validated in simulation and compared to a baseline MPCC without any CBF constraints. The case study presented in this paper considers a unicycle model, extended to include MPCC states as in \eqref{eq:unicycle_dynamics}. Obstacle detection is handled by a costmap $\mathcal{X}_O(t_k)$, updated at every timestep using measurements from a $360^\circ$ LiDAR. The receding horizon motion planner used in this work is an augmented version of the FASTER solver \cite{mohammad2024gp}, \cite{tordesillas2022faster}. Both MPCC implementations run at 10Hz and were written in C++ using ACADOS \cite{Verschueren2021}.


In the baseline controller (MPCC-BASE), the formulation follows \eqref{eq:mpcc_ocp} but excludes any CBF constraints like \eqref{eq:mpcc_cbf_const}. Meanwhile, the CBF-SAC-enabled framework (MPCC-CBF-SAC) incorporates the CBF constraints as formulated in \eqref{eq:corridor_cbf_abv}-\eqref{eq:corridor_cbf_blw}.  The SAC policy $\pi_\sigma$ was trained offline on $35$ worlds of varying obstacle densities, sampled from the BARN dataset \cite{perille2020barn}. The adaptation bounds for $\alpha$ were selected based on empirical analysis: $\alpha^-=.05$ led to overly conservative motion, while $\alpha^+=.25$ allowed for more aggressive motion while still enhancing safety. We set $\alpha_0=.25$ to allow the system to start with aggressive motion and adapt as necessary. To ensure the exclusion of obstacles from $\mathcal{C}$, we used the following parameter values for calculating $\Delta\xi$ as in \eqref{eq:sampling_condition}: the maximum raycast distance $d^+=.35$m, the costmap resolution $r_c=.05$m, and $\kappa^+=6$, which was found empirically. With these values, $\Delta\xi$ was set to $.016$m. In the motion planner, we set $s_r \leq 4$, limiting the number of samples to $250$, which is low enough to allow $\rupper(\cdot)$ and $\rlower(\cdot)$ to be computed at every control cycle. 

With the parameters defined, the frameworks were then tested in $45$ worlds, running $5$ trials per world, resulting in $225$ total testing simulations per framework. MPCC-BASE succeeded in $172$ trials, achieving a $76\%$ success rate, while MPCC-CBF-SAC with $\alpha$ adaptation succeeded in $200$ trials, reaching $89\%$. The $25$ failures in the full approach stem from challenges in generating the offset curves $\rupper(\cdot)$ and $\rlower(\cdot)$. When $\bm{r}(\cdot)$ has high curvature (e.g., during sharp, acute-angle turns), the corridor generation technique \cite{arrizabalaga2024corridor} can produce self-intersecting curves, leading to infeasibility in the MPCC OCP and, consequently, collisions. Robust offset curve generation for spline trajectories remains an open problem due to the lack of a closed-form solution \cite{jiulong2013offset} and is a potential avenue for future work. 

To illustrate the performance improvements of MPCC-CBF-SAC over MPCC-BASE, consider the scenario in Fig.~\ref{fig:simulation_results}. The system navigates from the start state $\bm{x}_0$ to a final goal $\bm{x}_g$. Since the motion planner optimizes for travel time, the planned trajectory $\bm{r}(\cdot)$ at timestep $k$ cuts too close to an obstacle (black in the costmap), leading to a collision because MPCC-BASE faithfully tracks $\bm{r}(\cdot)$ without safety considerations (Fig.~\ref{fig:no_cbf_sim_result}). 
In contrast, Fig.~\ref{fig:sac_cbf_sim_result} shows MPCC-CBF-SAC successfully navigating through the same environment at multiple time points.

\textbf{Remark 3:} Although it may appear that an obstacle is protruding into the safety corridor at point C, the occupancy grid $\mathcal{X}_O(t_k)$ at that timestep actually resembled the one shown Fig.~\ref{fig:no_cbf_sim_result}. Part of the obstacle was occluding the rest from the system's viewpoint. However, as the system gathered more information as it rounded the turn, the safety corridor $\mathcal{C}$ updated accordingly.

At point B, the safety corridor generated by the offset curves $\rupper(\cdot)$ and $\rlower(\cdot)$ prevents the system from drifting too close to the problematic obstacle. By point C -- where MPCC-BASE collided -- MPCC-CBF-SAC maintains a safe distance, ensuring the system remains within $\mathcal{C}$. 

At point C, $\overline{h}(\bm{x})$ dips close to $0$, indicating the system is approaching the boundary of the safe set. However, it never crosses this boundary, demonstrating that the framework successfully maintains safety. This is primarily due to the behavior of $\underline{\alpha}$, as shown in Fig.~\ref{fig:alpha_blw_plt}, where it increases as the system moves through Point C. This adjustment allows the system to move closer to $\rlower(\cdot)$ in order to distance itself from the problematic obstacle. This adaptive behavior demonstrates that $\pi_\sigma$ has effectively learned to adjust the parameters at runtime to balance safety and feasibility. Additionally, we note that $\Delta\overline{\alpha}_k$ and $\Delta\underline{\alpha}_k$ remain within their bounds, indicating that $\pi_\sigma$ also respects the adaptation constraints.



\subsection{Experiment}
\begin{figure}[th!]
    \centering
    \includegraphics[width=0.48\textwidth]{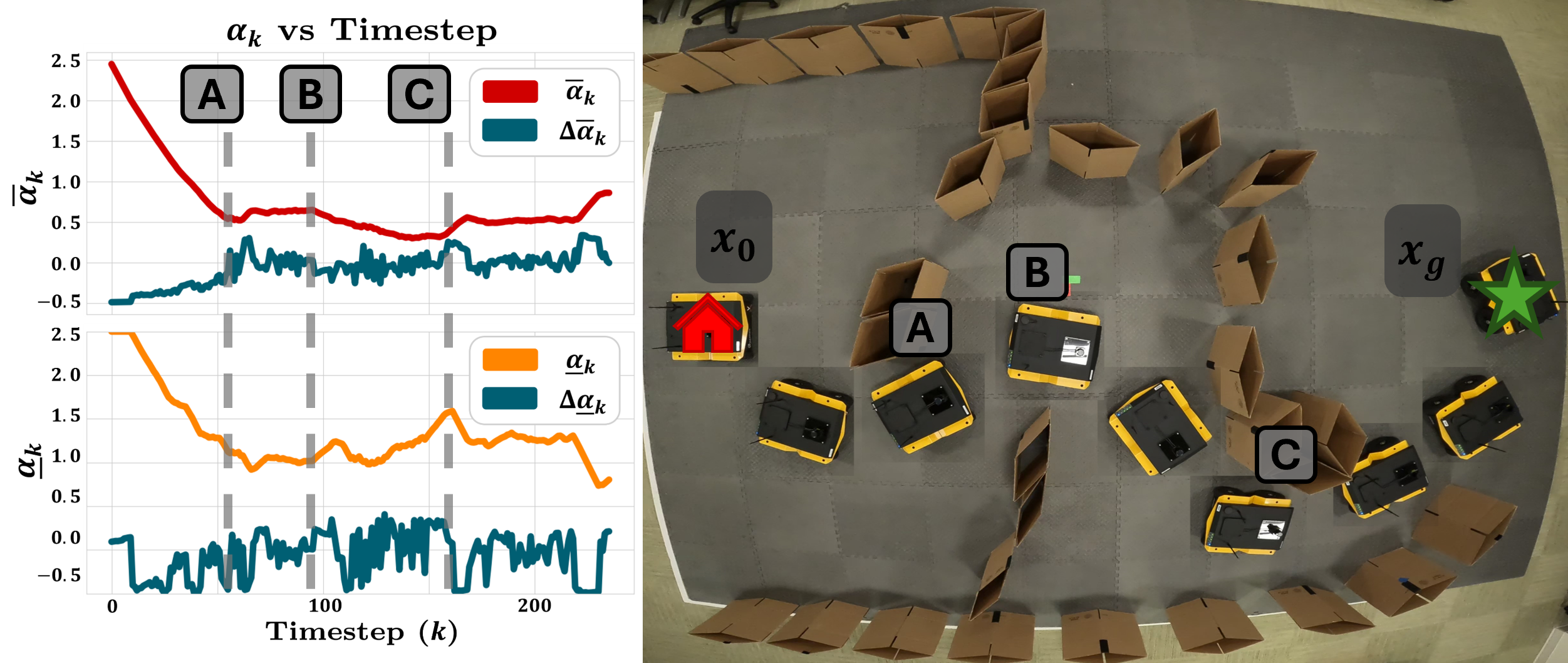}
    \caption{(Left) $\bm{\alpha}_k$ adaptation plots during navigation (Right) Jackal navigating through a cluttered environment using MPCC-CBF-SAC.}
    \label{fig:exp_figure}
    \vspace{-5pt}
\end{figure}

The proposed approach was further validated on a Clearpath Robotics Jackal, as shown in Fig.~\ref{fig:exp_figure}. Using the adaptation policy trained in simulation only, the MPCC-CBF-SAC was able to steer the jackal through the environment without colliding with any obstacles. The $\bm{\alpha}_k$ adaptation plots are also shown with key points labeled. Notice at point A that $\overline{\alpha}_k$ and $\underline{\alpha}_k$ both decrease as the system approaches the narrow corridor. $\overline{a}_k$ then increases while $\underline{\alpha}_k$ remains low at point B, pushing the system away from the protruding obstacles. At point C, $\overline{\alpha}_k$ steadily decreases while $\underline{\alpha}_k$ increases, pushing the system further towards the center of the corridor. 

\section{Conclusions and Future Work} \label{sec:conclusion} 
In this work, we have presented a novel CLF-CBF MPCC framework for general autonomous systems, leveraging receding horizon free space corridors to enforce CBF-based safety constraints for obstacle avoidance. We also extended our prior SAC-based CBF $\alpha$ parameter adaptation framework and demonstrated the effectiveness of our approach through simulations with a unicycle model case study.

Future work will focus on improving the robustness of the safety corridor generation, extending the approach to dynamic obstacles, and strengthening the theoretical guarantees for the feasibility of both the safety corridor CBF constraints and the $\alpha$ adaptation scheme. 



\bibliographystyle{IEEEtran}
\bibliography{library}

\end{document}